\documentclass[10pt,twocolumn,letterpaper]{article}

\usepackage{wacv}
\usepackage{times}
\usepackage{epsfig}
\usepackage{graphicx}
\usepackage{amsmath}
\usepackage{amssymb}

\wacvfinalcopy 

\def\wacvPaperID{105} 

\ifwacvfinal\fi
\begin{document}

\title{Video Generation from Text Employing Latent Path Construction for Temporal Modeling}

\author{Amir Mazaheri \hspace{3cm} Mubarak Shah \\
University of Central Florida - Center for research in Computer Vision (CRCV)\\
{\tt\small amirmazaheri@knights.ucf.edu \hspace{0.75cm} shah@crcv.ucf.edu}
}

\maketitle
\ifwacvfinal\thispagestyle{empty}\fi
\begin{abstract}
Video generation is one of the most challenging tasks in Machine Learning and Computer Vision fields of study. In this paper, we tackle the text to video generation problem, which is a conditional form of video generation. Humans can listen/read natural language sentences, and can imagine or visualize what is being described; 
therefore, we believe that video generation from natural language sentences will have important impact on  Artificial Intelligence.  Video generation is relatively a new field of study in Computer Vision, which  is far from being solved. The majority of recent works deal with synthetic datasets or real datasets with very limited types of objects, scenes and motions. To the best of our knowledge, this is the very first work on the text (free-form sentences) to video generation on more realistic video datasets like Actor and Action Dataset (A2D) or UCF101. We tackle the complicated problem of video generation by regressing the latent representations of the first and last frames and employing a context-aware interpolation method to build the latent representations of in-between frames.  We propose a stacking ``upPooling'' block to sequentially generate  RGB frames out of each latent representations and progressively increase the resolution. Moreover, our proposed Discriminator encodes videos based on single and multiple frames. We provide quantitative and qualitative results to support our arguments and show the superiority of our method over well-known baselines like Recurrent Neural Network (RNN) and Deconvolution (as known as Convolutional Transpose) based video generation methods.
\end{abstract}
\section{Introduction}
\begin{figure}
    \centering
    \includegraphics[width=\linewidth]{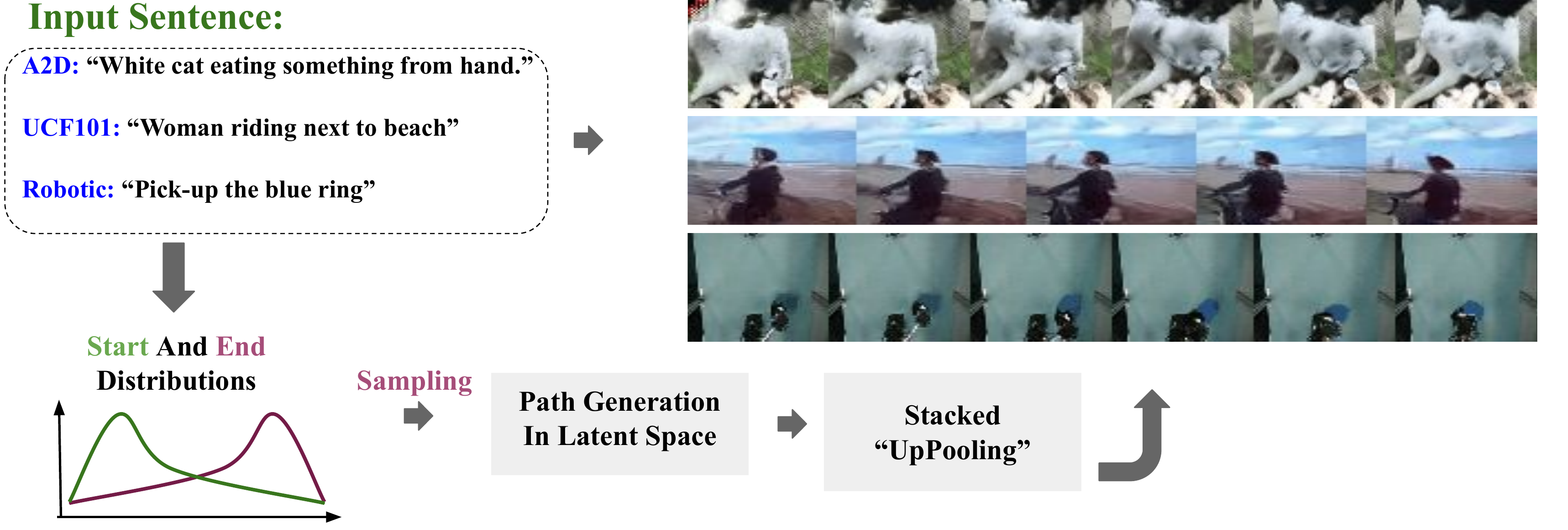}
    \caption{ Given an input sentence, we construct two distributions for the latent representations of the first and last frames. We build a path in the latent space between distributions of start and end frame. We generate high fidelity video frames by sampling from the latent constructed path through an ``UpPooling'' layer. }
    \label{fig:abstract}
\end{figure}

Videos and corresponding descriptions and  captions are  continuously being produced and stored every moment. This amount of joint video and text ``big data'' makes today the best time ever for Computer Vision and  Machine Learning (ML) to formulate and solve tasks related to a joint understanding of videos and text. In practice, Natural Language has been the primary tool for users to communicate with any video-based service. YouTube search query, textual descriptions as meta-data, textual comments for  expressing feelings, and IMDB (Internet Movie DataBase) blogs for summarization convince us that natural language is the best way for us to deal with video content.  We believe that similar to mentioned use cases, natural language is the best medium to create video content as well. It is easy for humans to express in detail what they want to see in the video, i.e.,  describing the colors, actions, objects. Plus, language is universal! Any human, with any background and skills, can express what he/she needs to create as a video! Considering all, we believe that generating videos from textual descriptions is a valuable task to study and solve, from both  computer vision  and real-world usability perspectives.

Realistically, a text to video content generating method must support free-form language and cover a broad set of activities, objects, etc. However, to the best of our knowledge, current works on text to video generation are mainly focused on synthetic datasets, or real datasets with a limited content domain, like only cooking videos. In this paper, we tackle the task of Video Generation with textual inputs for more realistic datasets i.e. videos in the wild; containing more natural videos and sentences compared to prior works. Please refer to  Figure~\ref{fig:abstract} for high-level overview of our approach.

A video generation model must be able to produce the spatial and temporal variations which have a natural coherence and consistency of a real video. Meanwhile, having a textual input sentence, while constraints the possible variations of a video, adds more complexity to the generative model, since the context of the input text must be visible and understandable from the generated video. Temporal dynamics is the key difference between videos and images, and makes the video generation a more challenging problem. Traditionally, Recurrent Neural Networks (RNNs) and Deconvolutions have been intuitive options to model the temporal dynamics of videos ; however, RNN and Deconvolution based methods add extra parameters to the model and contributes to the complexity of the network. Previous studies suffer from the lack of experiments to show if there is a need in the latent space to model the temporal dynamics with parametric components.  In this paper, we propose a novel method to capture temporal dynamics by first regressing the  first and last frames' latent representations from text description and employing a context-aware interpolation method to build up the latent representations of in-between frames. We show that our proposed simple but yet efficient model produces superior results compared to other known techniques like RNNs, or Deconvolution. 

\begin{figure*}[t]
    \centering
    \includegraphics[width = 1\textwidth]{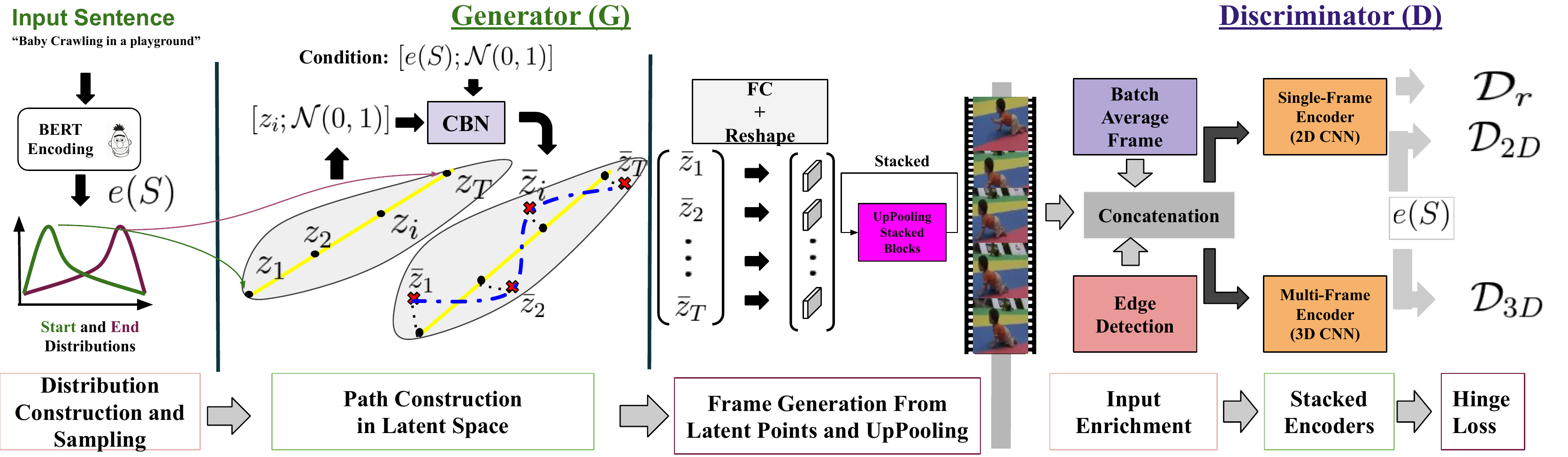}
    \caption{In this figure, we show a block-diagram of different steps of our proposed method. We encode the sentence using pre-trained BERT model and some trainable layers, and represent it by $e(S)$ (see Section~\ref{seq:TextEncoder} for details). Given $e(S)$, we construct two distributions and draw one sample from each corresponding to latent representations of start ($z_1$) and end ($z_T$) frames, respectively. We then determine T latent representations, [$\bar{z_1}, \bar{z_2},\dots,\bar{z_T}$], corresponding to $T$ frames,  employing a context-aware interpolation in the latent space. We use Conditional Batch-Normalization (CBN, Section~\ref{seq:batchnormalization}) with $e(S)$ and noise as the condition. Subsequently, we transform each $\bar{z}_i$ into a spatial representation using an FC and reshape layers, and increase its size to the desired resolution through stacked ``UpPooling'' blocks (Section~\ref{seq:FrameGenerator}). Inputs to the Discriminator are encoded sentence $e(S)$, and the video (real or fake). We augment the video input to the Discriminator by concatenating it with an average frame from the whole batch and edge maps of the frames. The discriminator employs a single and multi-frame based videos encoders along with $e(S)$, to measure if each frame and the video ($\mathcal{D}_{2D}$ and $\mathcal{D}_{3D}$) are relevant to the input sentence and if each spatial regions of each frame are naturally looking ($\mathcal{D}_r$). Finally, we train the proposed network with GAN Hinge-loss (Equations~\ref{eq:D_loss} and~\ref{eq:G_loss}).}
    \label{fig:BlockDiagram}
\end{figure*}

Our proposed method can generate high-fidelity videos for complicated datasets containing various kinds of actions, actors, camera view-points, etc.  We provide quantitative
results on  three different datasets: Actor-Action Dataset (A2D),  nine classes of UCF101 and Robot dataset with textual human-to-robot commands. 
To the best of our knowledge, we are the first one to solve the text to video generation problem on such challenging datasets like A2D, and UCF101, and report a comprehensive quantitative and qualitative study.


\section{Related Works}
\paragraph{\textbf{Video Generation:}}
Modeling the temporal coherency and consistency of natural video frames makes the video synthesis problem as one of the most challenging generative problems. Several types of the video synthesis have been studied in recent years. Authors in~\cite{chan2018everybody} solve the task of video synthesis as a conditional generation problem where the condition is the human pose skeleton; however, the method strongly depends on the human pose estimator and needs separate training data for each of human subjects, and scenes. Similarly, authors in ~\cite{siarohin2019animating} animate any arbitrary object given a driving video sample. This method detects a few key-points in each frame, and estimates a dense warping map by generalizing the key-points motion to all similar points.

Video synthesis can be combined with other computer vision tasks, like object~\cite{spampinato2018vos} or semantic segmentation~\cite{Pan_2019_CVPR}, Robot manipulation~\cite{finn2016unsupervised}, and etc. Authors in~\cite{spampinato2018vos} utilize video synthesis as an unsupervised technique to learn rich features for the task of Video Object Segmentation (VOS) with limited supervised data. They train a video generator by decomposing any video into a foreground object and a background scene. Similarly, authors in ~\cite{finn2016unsupervised} learn unsupervised features for robotic object manipulation tasks. Also, the work proposed in~\cite{Pan_2019_CVPR} generates videos conditioned on the first (only-first) semantically segmented frame. Similarly, authors in ~\cite{wang2018video} can generate videos out of a sequence of semantically segmented input frames.

Video generation can be also in form of video prediction~\cite{finn2016unsupervised,lee2018savp,lotter2016deep,villegas2017learning}, inpainting~\cite{kim2019deep}, etc.  Video prediction is to estimate the future frames of a video given the preceding ones. Video prediction is the most established and popular kind of video generation. The video inpainting task~\cite{kim2019deep}, similar to image inpainting~\cite{yu2018generative}, is to modify a specific spatial region of every single frame in a video.

A simplified form of video generation problem is to generate a video given a class label. 
Authors in~\cite{clark2019efficient} show that it is possible to generate high fidelity videos on a large number of classes. Similarly, the proposed method in~\cite{tulyakov2018mocogan} decomposes a video into content and motion subspaces and generates a video by sampling a point and a path in the two subspaces, respectively.

\paragraph{\textbf{Generation by Textual Input:}}
Textual sentences are the simplest form of natural human language; and transforming them into other mediums like video~\cite{pan2017create,Marwah_2017_ICCV,li2018video}, image~\cite{xu2018attngan,hinz2019semantic}, or speech~\cite{arik2017deep,huang1996whistler} is one of the most interesting problems in Artificial Intelligence. Authors in~\cite{xu2018attngan} propose a progressive~\cite{karras2017progressive} text to image generation method which leverages text to image attention at  multiple resolutions. 
Authors in~\cite{li2018video}, crawl YouTube by some selected search queries, and clean results to obtain a dataset for training the text to video neural network that produces a gist image from a sentence, and animate the gist image. However, sentences in~\cite{li2018video} are mostly in the form of ``Action'' + ``Place'', which is a simple-form compared to the sentences of our target dataset, A2D~\cite{gavrilyuk2018actor}. In this work, we use videos in the wild datasets like A2D~\cite{XuHsXiCVPR2015,gavrilyuk2018actor} and UCF101~\cite{soomro2012ucf101} (We provide the sentence annotations for nine classes of UCF101 in this paper). Datasets of our interest are not curated for the task of text to video generation and have complicated sentence structures. Authors in~\cite{Marwah_2017_ICCV} solve the task of video generation using text for simpler datasets like MNIST moving digits~\cite{srivastava2015unsupervised} and KTH action~\cite{schuldt2004recognizing}, using a Negative-Log-Likelihood (NLL) loss. 3D Deconvolutions and LSTMs have been used in~\cite{pan2017create} and ~\cite{liu2019cross} to generate multiple frames to generate a video. In this work, we propose our novel method to generate any number of needed frames to synthesis a video, and we show the performance of text to video generation on more challenging datasets.

\section{Approach}

Our proposed method to solve the text to video generation follows the Generative-Adversarial framework, which includes a generator and a discriminator sub-modules. In Figure~\ref{fig:BlockDiagram}, we show  our method diagram, including all the steps in both Generator(G) and Discriminator(D). In the rest of this section, we present  the details of our proposed model.

\subsection{Text Encoder}\label{seq:TextEncoder}
Given a sentence as the sequence of words $S = [w_1, w_2, \dots, w_N]$, the purpose of the text encoder is to represent $S$ as a vector of real numbers. Ideally, one can train a neural network from scratch or end-to-end integrate with the rest of the system, similar to~\cite{pan2017create}. However, in this paper we target  realistic datasets, i.e., A2D and UCF101,  and due to the complex nature of such target datasets and annotations, we do not have enough number of examples for each of the words in the dataset.  A large portion of the words in our target datasets are rare words; moreover, there are many words in the test set, which are not seen  during the training. For example, more than 500 verbs, adjectives, and nouns in the A2D dataset appear only once. Also, this amount of rare words makes models like~\cite{xu2018attngan} impractical.

We employ the BERT (Bidirectional Encoder Representations from Transformers)~\cite{devlin2018bert} sentence encoder, pre-trained on English Wikipedia\footnote{https://github.com/hanxiao/bert-as-service}. BERT provides us a rich representation of all the sentences even if they contain rare words. We transform the $1024$ dimensional output of BERT encoding into $256$ D using two blocks of Fully-Connected, Batch-normalization, and Leaky-ReLU layers. We denote the encoded sentence by $e(S) \in \mathcal{R}^{256}$.
\subsection{Video Generator}
Let $e(S)$ be the encoded sentence. We estimate two Gaussian distributions $\mathcal{N}_s (\mu_s, \sigma_s) \in \mathcal{R}^{256}$, and $\mathcal{N}_e (\mu_e, \sigma_e) \in \mathcal{R}^{256}$ for the starting and ending frames:

\begin{equation}
    \mu_s, \mu_e, \sigma_s, \sigma_e = \mathcal{F}([e(S) ; \mathcal{N}(0, 1)]).
\end{equation}
Here, $\mathcal{F}$ is a Multilayer perceptron (MLP). Concretely, we split the output of the $\mathcal{F}$ into four equally length vectors, and we use sigmoid non-linearity on top $\sigma_s$ and $\sigma_e$. Note that, $[;]$ denotes concatenation operation throughout this manuscript. We draw one vector from each of the distributions $\mathcal{N}(\mu_s, \sigma_s)$ and $\mathcal{N}(\mu_e, \sigma_e)$, and denote them by $z_{1}$ and $z_{T}$. To generate a video with $T$ frames, we employ an interpolation to extract the latent representation for frame $i$:
\begin{equation}
    z_i = \dfrac{{T - i}}{T} z_{1} + \dfrac{i}{T}  z_{T}.
\end{equation}
We choose the linear interpolation for this step as the most simple option. Our observations show that more complicated interpolations like bi-cubic or spherical linear interpolation (SLERP) are not as good as linear interpolation in latent space.  We concatenate each of $z_i$ vectors with a normal noise vector $\mathcal{N}{(0,1)} \in \mathcal{R}^{32}$ and pass them through a Conditional Batch-Normalization (CBN)~\cite{de2017modulating} (see Section~\ref{seq:batchnormalization}), where the condition is $[e(S);\mathcal{N}(0,1)]$.  Normalized latent representations are denoted by $\bar{z_i}$.  The CBN module and its effect  on the training are briefly explained in Section~\ref{seq:batchnormalization}. The added random noise  $\mathcal{N}(0,1)$ brings in the needed variability to the final motion. In addition, the CBN provides a stochastic context-aware transformation on each latent representation, to finally produce $\bar{z}_1 \dots \bar{z}_T$.%


\subsubsection{Frame Generator} \label{seq:FrameGenerator}
\begin{figure}
    \centering
    \includegraphics[width=0.7\linewidth]{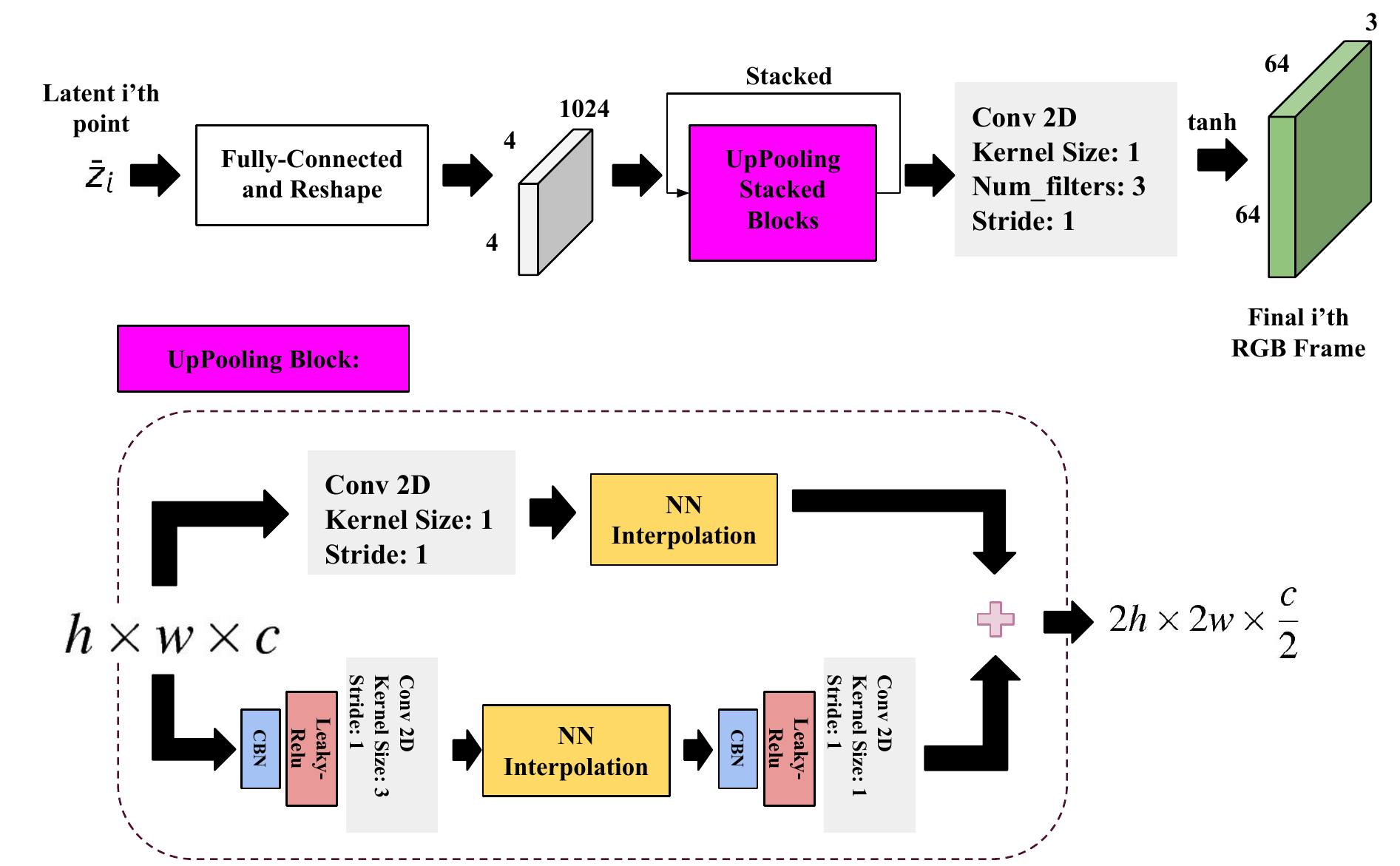}
    \caption{Here we present the frame generator sub-module. We transform each latent vector,  $\bar{z}_i$, which corresponds to i'th frame of the video, into a spatial representation using a Fully-Connected layer and reshape. We increase the resolution of the spatial representation by a factor of two in each ``UpPooling Block''. The UpPooling block has a short and a long path. The short path has only a linear $1\times1$ Convolution and the long path, which has two 2D CNNs along with Conditional Batch Normalization (CBN) and Leaky-ReLU. We increase the resolution in both paths by a Nearest Neighbour(NN) interpolation. Note that,  the concatenation of the encoded text and a $32$ dimensional noise ($[e(S);\mathcal{N}(0,1)]$) are used as a condition to all CBNs. Finally, we build the RGB frame by a $1\times1$ Convolution and tanh non-linearity.}
    \label{fig:upPooling}
\end{figure}

In the second part of our Generator network, we propose a CNN based network to transform each of $z_{i}$ latent vectors into a RGB frame. First, the latent vectors are transformed  into a spatial representation using a linear transformation and reshape. Basically,  each  $\bar{z}_{i}$ is mapped into a $h_1 \times w_1 \times c_1$ vector using a Fully-Connected layers, and are reshaped  into a spatial tensor $ \in \mathcal{R}^{h_1\times w_1\times c_1}$. In our experiments, $h_1=4$, $w_1=4$, and $c_1 =  2048$ are employed.

To build the frames of desired resolution,  a CNN based module is employed to increase (up-pooling) the resolution of spatial features (see Figure~\ref{fig:upPooling}). The proposed module increases the resolution of the given input via two paths, a short path with only one convolution layer, and a longer path with two convolution layers with Conditional Batch-Normalization (Section~\ref{seq:batchnormalization}) and ReLU activation in between . The short path plays a role of skip-connection that facilitates the training process. However, the longer path increases the capacity of the model by adding non-linearity and normalization.   Nearest-Neighbour (NN) interpolation is used  to increase the spatial size of each tensor. We tried  PixelShuffle~\cite{shi2016real}, and 2D-Deconvolutions as other design choices, however, NN-interpolation consistently produced  better results in all experiments.

We stack the ``UpPooling'' block (as explained in Figure~\ref{fig:upPooling}) to reach the desired output resolution. In our experiments, our generated frames are $64 \times 64$; thus, we need four blocks of UpPooling Blocks.
Finally, we apply a 3D convolution on the output of the final layer with 3 (RGB) filters and $tanh$ non-linearity to build the final RGB frame.

\subsection{Conditional Batch-Normalization (CBN)} \label{seq:batchnormalization}
Here, we briefly explain the conditional Batch-normalization we employ in our generator. Given an input $x$ and condition $c$ we compute $\bar{x}$ as follows:
\begin{equation}
    \bar{x} = {{\mathcal{\gamma}}(c)}_{\mu = 1}\frac{x - \mu_{x}}{\sigma_{x}} + \mathcal{\beta}(c)_{\mu = 0},
\end{equation}
where $\mathcal{\gamma}(.)$ and $\mathcal{\beta}(.)$ are neural networks that have the same output shape as shape of $x$. In our case, we use a single linear FC layer to implement each of them. Also, $\gamma(c)_{\mu = b} = \gamma(c) - \mu_{\gamma(c)} + b$ (same for $\beta(c)$, that simply means to shift the mean value of the batch, $\mu_{\gamma(c)}$, to $b$), and we compute the mean $\mu$, and variance $\sigma$ over the batch. Conditional Batch-Normalization, in fact, normalizes the mean and variance of each sample with respect to the statistical data of whole batch, and applies a context-aware affine transformation (scale ${\mathcal{\gamma}}(c)$, and shift $\mathcal{\beta}(c)$), where the context is represented as condition $c$, on the normalized input.

\subsection{Discriminator}
\begin{figure}
    \centering
    \includegraphics[width=0.7\linewidth]{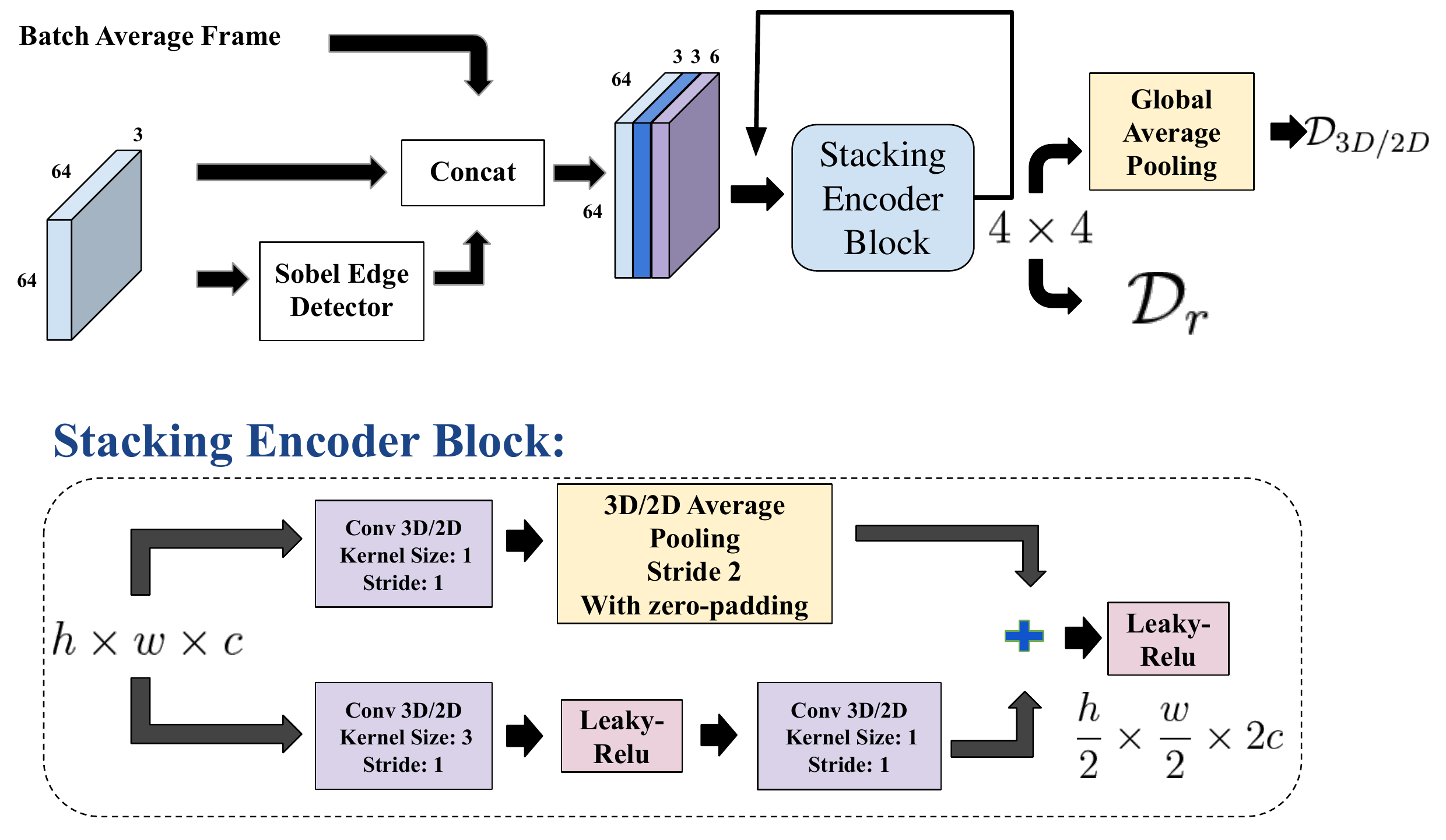}
    \caption{Our proposed Discriminator building block. First, we compute the average frame of a batch and each frames' edge map using the Sobel edge detector and concatenate them to the input frames. To reduce the input resolution, we implement a stacking encoder block that consists of a short path of a $1\times1$ convolution followed by average pooling, and in parallel, a long path with a $3 \times 3$ convolution followed by average pooling and a $1 \times 1$ Convolution. We sum the outputs of short and long paths, which have half of the resolution of the input. We stack this block until we reach a  $4 \times 4$ spatial resolution. Note that, for the multi/single-frame based discriminator, we use 3D/2D convolutions and average pooling layers.}
    \label{fig:DiscBlock}
\end{figure}
Our proposed Discriminator (D) consists of a frame-based and a  video-based encoder sub-modules.  The frame-based sub-module encodes each frame globally and locally using 2D CNN blocks. It encodes each frame into one vector (global frame encoding), and estimates its relevance to the input text, while it uses spatial features, extracted before the global average pooling (see Figure~\ref{fig:DiscBlock}), to compute one score for each region of the frame. It helps the discriminator not only to determine if the global context of the video is related to the text, but also each spatial region of the frames is locally natural looking. Similarly, the multi-frame (video-based) sub-module D, leverages 3D CNN blocks to encode all frames of a video, as a sequence, into a vector. To compute the relevance between the encoded video $v \in \mathcal{R}^{d_{v}}$, which can be a single or multi-frame based encoded vector,  and the encoded sentence $e(S) \in \mathcal{R}^{d_{e}}$, we compute the discriminator score by:
\begin{equation}
    \mathcal{D}(v, e(s)) = W_{D} \times ( \sigma(W_{e} \times e(S))\odot v),
\end{equation}
where $\odot$ represents element-wise multiplication, $W_{D} \in \mathcal{R}^{d_{v} \times 1}$ and $W_{e} \in \mathcal{R}^{d_{e} \times d_{v}}$. We denote the discriminator scores from 3D CNN multi-frame video-based encoder by $\mathcal{D}_{3D}$, and from 2D CNN frame-based encoder by $\mathcal{D}_{2D}$. Also, we use $\mathcal{D}_{r}$ for the spatial regions' scores which is computed along with the single-frame encoder. Note that, $\mathcal{D}_{r}$ is a function of single frames, and is independent of $e(S)$. Finally, we take an average of all the scores from all the frames to compute the final $\mathcal{D}_{2D}$ and $\mathcal{D}_{r}$.

\subsubsection{Discriminator Input Enrichment} \label{seq:DiscEnrich}
We observe that using Batch-Normalization or Conditional Batch-Normalization in Discriminator (D) architecture do not facilitate the training process. In our experiments, D containing BN dominates the G in early iterations, and results in severe mode-collapse. To utilize the stochastic batch information, we propose to concatenate each RGB frame  with an average RGB frame of all the frames in a batch. In this scenario, D benefits from information in both single sample and batch statistics. This technique reduced the mode-collapse in our experiments; without this technique, we observe that there is a high chance that the model collapses into one or two modes during training, and we need to reload an earlier checkpoint to continue training. Additionally, as shown in previous studies~\cite{fu2019edge}, edge information is essential to detect if a frame is blurred. We augment each RGB frame with its Sobel edge map. See Figure~\ref{fig:DiscBlock} for more details.

\subsection{Loss Function} \label{seq:LossFunction}
We use the hinge-based loss to train our GAN architecture. We compose the Generator loss as:

\begin{equation} \label{eq:G_loss}
    \mathcal{L}_G = -\mathcal{D}_{3D}(G(S), e(S)) -\mathcal{D}_{2D}(G(S), e(S)) - \mathcal{D}_{r}(G(S)),
\end{equation}
and the discriminator loss as:
\begin{multline} \label{eq:D_loss}
    \mathcal{L}_D = [1 - \mathcal{D}_{3D}(\mathcal{V}, e(s))]_{+} + [\mathcal{D}_{3D}(G(S), e(S)) + 1]_{+} + \\
    [1 - \mathcal{D}_{2D}(\mathcal{V}, e(S))]_{+} + [\mathcal{D}_{2D}(G(S), e(S)) + 1]_{+} + \\
    [1-\mathcal{D}_{r}(\mathcal{V})]_{+} + [\mathcal{D}_{r}(G(S))+1]_{+},
\end{multline}
where $[x]_{+} = max(0, x)$ and $\mathcal{V}$ is a real video from training set with the text annotation $S$, and $G(S)$ is a generated video given $S$.
\section{Experimental Results}

\subsection{Dataset}
\noindent \textbf{Actor and Action (A2D)}
The A2D dataset is a popular dataset for the actor and action video segmentation task. The authors in~\cite{gavrilyuk2018actor} provide sentence annotations corresponding to video actor segmentation, and authors in~\cite{mcintosh2018multi} provide frame level bounding box for each actor. A2D contains 3,782 videos with  6,656 sentences corresponding to actor segmentation, including 811 different nouns, 225 verbs, and 189 adjectives. Each sentence corresponds to one actor  throughout the whole video. Therefore, there can be more than one sentence annotated for each video, corresponding to multiple actors. We crop the video for each sentence by constructing a maximal bounding box that covers all the instances of the  object in all the frames. This way  we get  one video sequence for each sentence; hence,  6,656 cropped video sequences and sentences.\\
\textbf{UCF101} is one of the popular datasets for the task of human action recognition. However, to the best of our knowledge, there has been no video level captioning annotations for UCF101. We have annotated 9 classes of UCF101. The selected classes are: ``Fencing'', ``Basketball'', ``Basketball Dunk'', ``Biking'', ``Golf Swing'', ``Gymnastics'', ``Cricket Bowling'', and ``Cliff Diving''. We asked the annotators to describe each video by a short sentence. Note that some of videos in each of UCF101 classes are very similar, and we let the annotators use identical annotations based on their judgment. The corpus of video captions have 182 unique words, and the maximum sentence length is 22 words.\\
\textbf{Robotic Videos:} Authors in~\cite{abolghasemi2019pay} provide an object manipulation robotic dataset containing videos and corresponding user-to-robot textual command. This dataset contains ``push'' and ``pick-up'' tasks for multiple objects. Sentences are in form of ``task + object description''. For example, ``pick-up the blue box''. Each video is about 20 seconds, and we randomly pick 16 frames to train the system.

\subsection{Evaluation Metrics}\label{seq:evaluationMetrics}
\noindent \textbf{Inception Score (IS) ~\cite{salimans2016improved}}is widely used in quality assessment of  generative models. Inception Score is computed based on a pre-trained classifier on the dataset. Ultimately, any generated sample must belong to a specific class (high probability output on a single activation of the classifier), and the model must generate outputs from all the available categories (diversity on the classifier output). A higher IS is better.
To compute the Inception Score, we fine-tune the I3D model~\cite{carreira2017quo}  pre-trained on Kinetics~\cite{kay2017kinetics} and imagenet~\cite{deng2009imagenet}  on each of datasets with the same number of classes (in our case, 43,9 and 11 classes for A2D, UCF101, and Robotic datasets, respectively), and other settings like the frame size, frame rate (fps), etc.
We fine-tune the pre-trained  I3D model~\cite{carreira2017quo} on Kinetics~\cite{kay2017kinetics} and imagenet~\cite{deng2009imagenet}.\\
\textbf{Fréchet Inception Distance (FID)}~\cite{heusel2017gans} compares the statistics of two sets of samples, namely real and fake. We use the same fine-tuned I3D classifier used for the IS score and extract $1024$ dimensional features. The lower FID is better. This quantitative measurement for video synthesis is also known as FVD~\cite{clark2019efficient}.\\
\textbf{R-Precision:}Following~\cite{xu2018attngan}, we employ R-Precision, which is a retrieval evaluation metric. To compute the R-Precision, we first train a  CNN based retrieval network (again based on pre-trained I3D), that can rank a set of sentences, given a video. The retrieval network architecture consists of a video encoder and a text encoder, and we train it with a hinge ranking loss and cosine similarity until it fully converges on the training data. This network achieves  ``top-1 accuracy'' of 80\% and 60\% for UCF101 and A2D training data, respectively. Later, given a sentence, we generate a video and using the retrieval network, we rank a set of 100 sentences, including unseen and seen. Assuming that there are $R$ related sentences in the 100 sentences, and $r$ of them are in top $R$ ranked sentences, the R-precision score is: $\dfrac{r}{R}$.
Note that, in contrast to~\cite{xu2018attngan}, our datasets do not have multiple sentences per video sequence. Or simply, $R=1$. To overcome this issue, we slightly alter each sentence by randomly dropping/replacing some words. Using this technique, we generate between 6 to 12 related sentences for each video. We believe that if a sentence is slightly changed, it must be still ranked above totally unrelated sentences. We use this metric for A2D and UCF101 datasets.
\textbf{Accuracy}: Since there are only 11 unique sentence in the Robotic dataset~\cite{abolghasemi2019pay}, and some of them have only one word difference, the R-Precision is not a good option for evaluation. Instead, we train a classification network that given a video, classifies which of the 11 classes (unique sentence) the video belongs to. Later, we use this classification network and test it on the generated videos. A higher accuracy is better.

\begin{table}[]
    \centering
    \resizebox{\linewidth}{!}{%
    \begin{tabular}{|c|c|c|c|c|}
        
        \hline
         & IS $\uparrow$ & All-FID $\downarrow$ & Intra-FID $\downarrow$ & R-P$\uparrow$ \\
         \hline
        \multicolumn{5}{c}{Temporal Modeling Baselines}\\
         \hline
        
        Deconvolution~\cite{pan2017create} & $3.84 \pm 0.12$  & 31.56 & 108.04 & 0.31 \\
        \hline
        SLERP + LSTM~\cite{spampinato2018vos} & $4.04 \pm 0.17$ & {\bf 25.74} & 104.13 & 0.34 \\
        \hline
        ConvGRU~\cite{clark2019efficient} & $3.97 \pm 0.33$ & 34.78 & 109.78 & 0.30 \\
        \hline
        \multicolumn{5}{c}{Ablation Study on Textual Encoding}\\
         \hline
          Only Class Labels & $3.90 \pm 0.12$ & 49.35 & 119.37 & N/A \\
        \hline
        BiLSTM Sentence Encoder & $3.09 \pm 0.13$ & 54.12 & 131.58 & 0.05 \\
        \hline
        \multicolumn{5}{c}{Ablation Study on Discriminator}\\
         \hline
         Frame Based Discriminator ($\mathcal{D}_{2D})$ & ${ 3.52} \pm { 0.11}$  & 169.59 & 49.17 & 0.06 \\
         \hline
         Video Based Discriminator ($\mathcal{D}_{3D}$) & ${ 3.17} \pm { 0.09}$ & 39.75 & 117.92 & 0.08 \\
         \hline 
         Region-Based Discriminator ($\mathcal{D}_{r}$) & ${ 3.77} \pm { 0.20}$ & 38.14 & 100.18 & 0.05\\
         \hline
         \hline
         
        \textbf{Ours - Full Model} & ${\bf 4.85} \pm {\bf 0.16}$ &  25.91  & {\bf 94.26} & {\bf 0.39} \\
        \hline
        \hline
        \textit{Real Data} & $\textit{9.93} \pm \textit{1.18}$ & \textit{11.55} & \textit{71.56} & \textit{0.45} \\
        \hline
    \end{tabular}
    }
    \caption{A2D experimental results. All-FID: The FID on all the videos from all classes. Intra-FID: Mean of FID within videos of each class. R-P: R-Precision. IS: Inception Score. }
    \label{tab:A2D}
\end{table}

\subsection{Quantitative Results}
We evaluate our trained model on UCF101 and A2D dataset using the explained metrics in Section~\ref{seq:evaluationMetrics}. For a comprehensive study, we include some baselines in which we use other design choices from previous works. We provide the following baselines for all  A2D, UCF101, and Robotic datasets.Note that the original implementations of baselines are not designed for this dataset and problem, i.e. text to video generation. Hence, to have a fair comparison, we use our best text encoding and discriminator for each baseline. Also, we carefully tune the hyperparameters for each baseline.
\textbf{Only Class Labels:} We train the model merely with video class labels. By comparing the results of this method with our final method, we show that sentences are more compelling conditions for the task of video generations; and our generative model benefits from additional information contained in a sentence compared to employing only labels.\\
\textbf{SLERP + LSTM:} We follow the design of~\cite{spampinato2018vos} which construct the temporal domain of a video by a Spherical Linear Interpolation (SLERP), and estimates each latent point representation $\bar{z}_i$ using an LSTM.\\
\textbf{Deconvolution:} In this baseline, similar to~\cite{pan2017create}, we expand the number of generated frames by Stacking Deconvolution layers (also known as Convolution Transpose).\\
\textbf{Conv-RNN:} Similar to~\cite{clark2019efficient}, we estimate a distribution out of the input text, and transform it into a spatial representation using a linear transformation and reshape. The resulting spatial representation is repeated $T$ times and is passed to a Convolutional Recurrent Neural Network. We observe that a Convolutional Gated Recurrent Unit (ConvGRU) with layer Normalization is the best choice for this baseline.\\
\textbf{Real Data:} Evaluation on the ``Real Data'' gives us a better understanding of what would be a realistic expected value for each of the Inception Score, Fréchet Inception Distance, and R-Precision. We do not expect even on real data, we will get the best possible scores, since neither of I3D or our retrieval network is perfect. Note that, the FID value would be ideally zero on the real data itself; however, we split the set of all the real videos in half and compute the FID between these two sets.

\begin{table*}[t]
    \centering
    \resizebox{0.93\textwidth}{!}{%
    \begin{tabular}{|c|c|c|c||c|c|c|c|c|c|c|c|c|c|}
        \hline
        \multicolumn{4}{|c||}{
        All-Videos}& \multicolumn{10}{c|}{Intra-Classes FID}\\
        \hline
        
         & IS $\uparrow$ & R-P $\uparrow$ & FID $\downarrow$ & \rotatebox{90}{\small Fencing} & \rotatebox{90}{\small Basketball}& \rotatebox{90}{\small B. Dunk}& \rotatebox{90}{\small Biking}& \rotatebox{90}{\small Diving}& \rotatebox{90}{\small Golf}& \rotatebox{90}{\small Gymnastics}& \rotatebox{90}{\small Crick. Bowl.}& \rotatebox{90}{\small CliffDiving}& \rotatebox{90}{\textit{Mean}} \\
         \hline
         Only Class Labels & $3.69 \pm 0.19$ & N/A & 60.38 & 163.97 & 77.93 & 131.31 & 144.57 & 49.37 & 48.52 & 121.26 & 77.89 & 39.86 & 94.96\\
        \hline
        SLERP + LSTM~\cite{spampinato2018vos} & $1.02 \pm 0.00$ &  0.04 & 127.05 & 181.59 & 132.14  & 132.03 & 190.87 & 222.27 & 101.90 & 134.91 & 174.22 & 94.22 & 151.57\\
        \hline
        Deconvolution~\cite{pan2017create} & $3.95 \pm 0.19$ & 0.19 & 51.64  & 126.85 & 116.87 & 53.06 & 98.28 & 85.80 & 59.81 & 105.05 & 103.95 & 49.91  & 88.84\\
        \hline
        ConvGRU~\cite{clark2019efficient} & $5.93 \pm 0.18$ & 0.35 & 30.24 & 63.31 & 54.99 & 70.03 & 66.61 & 68.52 & 23.01 & 90.65 & 40.53 & 35.89 & 57.06 \\
        \hline
        \textbf{Ours} & ${\bf 7.01} \pm {\bf 0.36}$ & {\bf 0.43} & {\bf 17.12} & {\bf 29.20} & {\bf 28.08} & {\bf 54.69} & {\bf 46.48} & {\bf 48.54} & {\bf 19.44} & {\bf 46.24} & {\bf 31.40} & {\bf 35.44} & {\bf 37.72}\\
        \hline
        \hline
        \textit {Real Data} & $\textit{8.24} \pm \textit{ 0.20}$ & \textit{0.56} & \textit{6.92} & \textit{14.51} &  \textit{18.71} & \textit{16.21} & \textit{11.19} & \textit{16.76} & \textit{3.87} & \textit{18.75} & \textit{12.64} & \textit{15.78} & \textit{14.27}\\
        \hline
    \end{tabular}
    }
    \caption{UCF 101 Quantitative Results. Here we report the Inception Score (IS), R-Precision (R-P), Fréchet Inception Distance (FID). For the FID score,  all the videos from all the classes of the dataset, are used to compute the FID score. And in another experiment (Intra-Classes FID) we compute the FID score for the videos within each class.}
    \label{tab:UCF101}
\end{table*}

Furthermore, for the A2D dataset, which is the most challenging dataset, we provide more ablation studies (Table~\ref{tab:A2D}) to show the contribution of the proposed components in our method. In \textbf{BiLSTM Sentence Encoder} experiment, we replace the pre-trained BERT encoder with a simple BiLSTM that trains from the scratch, and the performance of the method drops drastically. This is due to the reasons mentioned in Section~\ref{seq:TextEncoder}. Moreover, we provide ablation study on the Discriminator. We isolate each of the discriminator terms, namely $\mathcal{D}_{r}$, $\mathcal{D}_{2D}$, and $\mathcal{D}_{3D}$. By comparing the performance of these ablation studies with the full model, we show that the terms in Equations~\ref{eq:G_loss}, and~\ref{eq:D_loss} are complementary.
\begin{table}[t]
    \centering
    \resizebox{\linewidth}{!}{%
    \begin{tabular}{|c|c|c|c|c|}
        \hline
         & IS $\uparrow$ & All-FID $\downarrow$ & Intra-FID $\downarrow$ & Accuracy ($\%$) $\uparrow$ \\
         \hline
         Only Class Labels & $1.99 \pm 0.19$ & 20.39 & 73.2 & 10.4\\
        \hline
        Deconvolution~\cite{pan2017create} & $2.97 \pm 0.21$  & 6.59 & 18.49 & 70.4 \\
        \hline
        SLERP + LSTM~\cite{spampinato2018vos} & ${\bf 3.47} \pm {\bf 0.16}$ &  4.60 & 18.22 & 73.7 \\
        \hline
        ConvGRU~\cite{clark2019efficient} & $3.17 \pm 0.26$ & 6.65 & 25.74 & 50.4 \\
        \hline
        \textbf{Ours} & $ 3.36 \pm  0.15$ &  {\bf 3.79}  & {\bf 16.45} & {\bf 76.6} \\
        \hline
        \hline
        \textit{Real Data} & $\textit{3.64} \pm \textit{0.3}$ & \textit{3.4} & \textit{11.8} & \textit{100}\\
        \hline
    \end{tabular}
    }
    \caption{Robotic experimental results. All-FID: The FID on all the videos from all classes. Intra-FID: Mean of FID within videos of each class.  IS: Inception Score.}
    \label{tab:Robotic}
\end{table}

For the sake of fairness, we keep the implementation of all the baselines and our proposed method as similar as possible. For example, the discriminator architecture, hardware, and etc.
In Tables~\ref{tab:A2D},~\ref{tab:UCF101}, and ~\ref{tab:Robotic} we show the results of our proposed method respectively on A2D, UCF101, and Robotic datasets. 
Our proposed method is competitive to the baselines based on all the evaluation metrics.


\subsubsection{Qualitative Results}
In Figures~\ref{fig:A2DQualitative} and~\ref{fig:UCF101Qualitative} we provide qualitative results for A2D and UCF101 datasets. Each figure comes with multiple sentences and generated videos corresponding to each of them. In Figure~\ref{fig:RobotQual}, we show generated videos that contain 16 frames. Note that, for the Robot dataset, each video represents a full task performance, which usually has around 200 frames in the original dataset. These results show that our method can handle datasets with a higher skip frame rate (lower fps). 
In more realistic and wild datasets like A2D, videos can have various ranges of motion. A video can have minimal motion (static video) or jumpy consecutive frames. We observe that our model can successfully cover various motions. For example, in Figure~\ref{fig:A2DQualitative}, the top left example (``The bird is climbing the chair'') has much less motion than the bottom left example (``A bird is flying'').

\subsection{Experimental Setup Details}
\begin{figure*}
    \centering
    \includegraphics[width=0.85\textwidth]{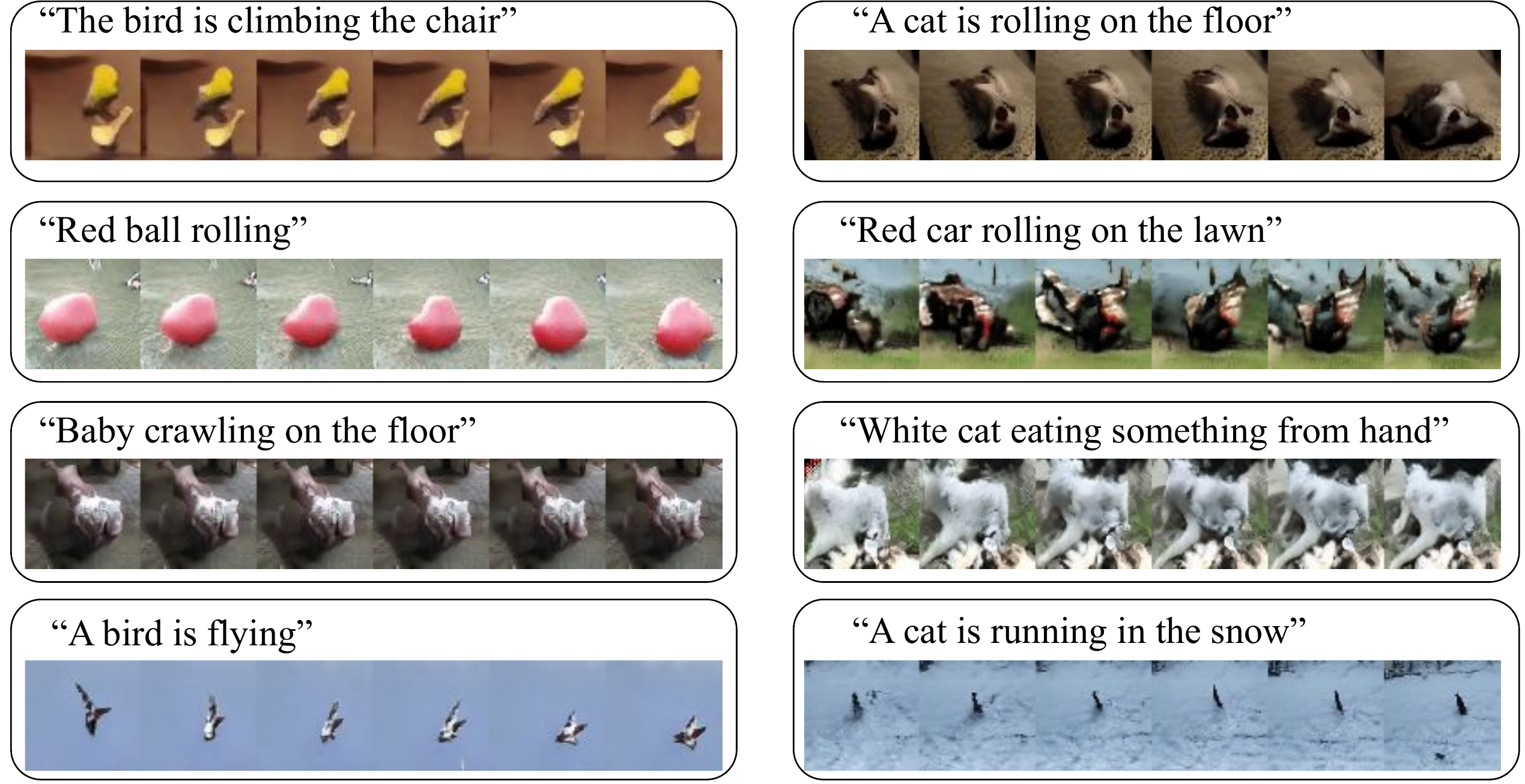}
    \caption{Qualitative Results on A2D dataset. Corresponding to each sentence we show the frames of generated videos. All samples are 6 frames with $64 \times 64$ resolution. Our proposed model can successfully produce diverse videos with different amount of motions, backgrounds and objects.}
    \label{fig:A2DQualitative}
\end{figure*}

\begin{figure*}
    \centering
    \includegraphics[width=0.85\textwidth]{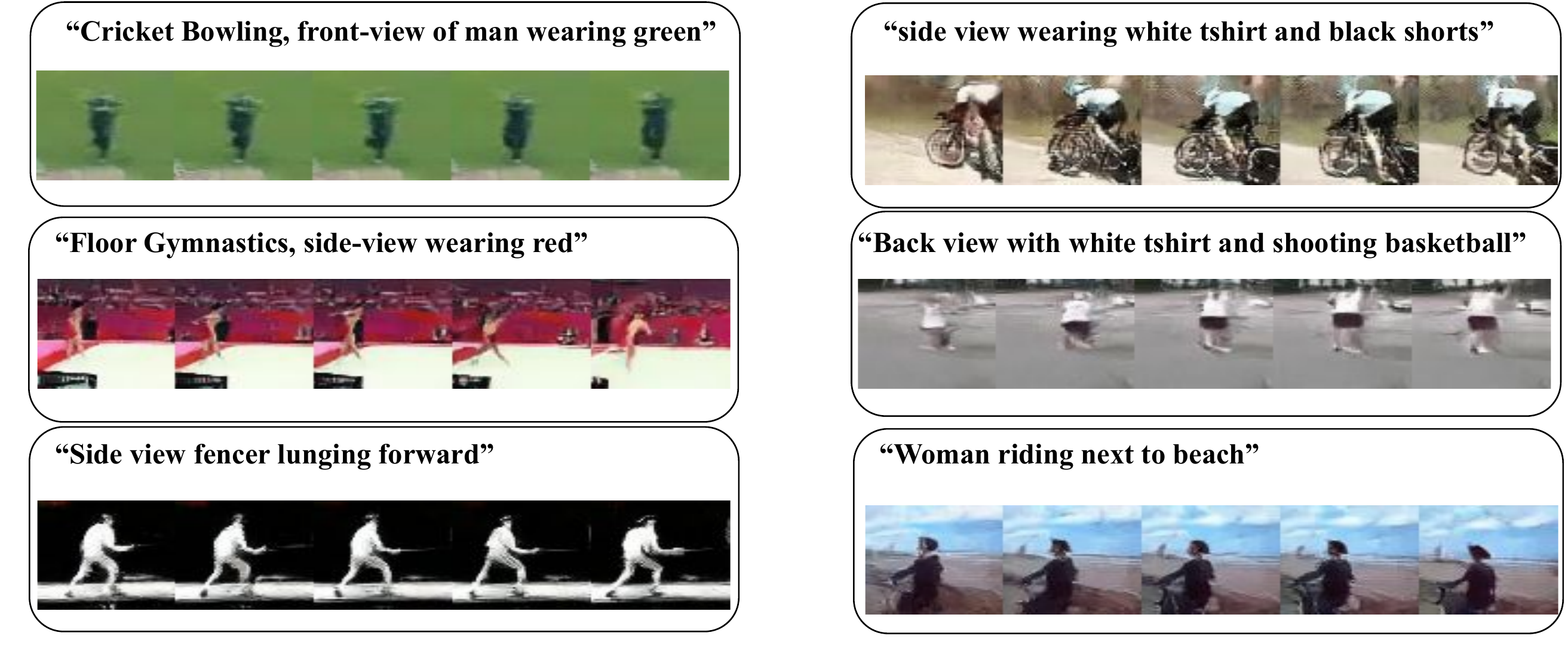}
    \caption{UCF101 qualitative results. Corresponding to each sentence we show the frames of generated videos. All samples are 6 frames with $64 \times 64$ resolution.
    }
    \label{fig:UCF101Qualitative}
\end{figure*}

\begin{figure*}[h!]
    \centering
    \includegraphics[width=\textwidth]{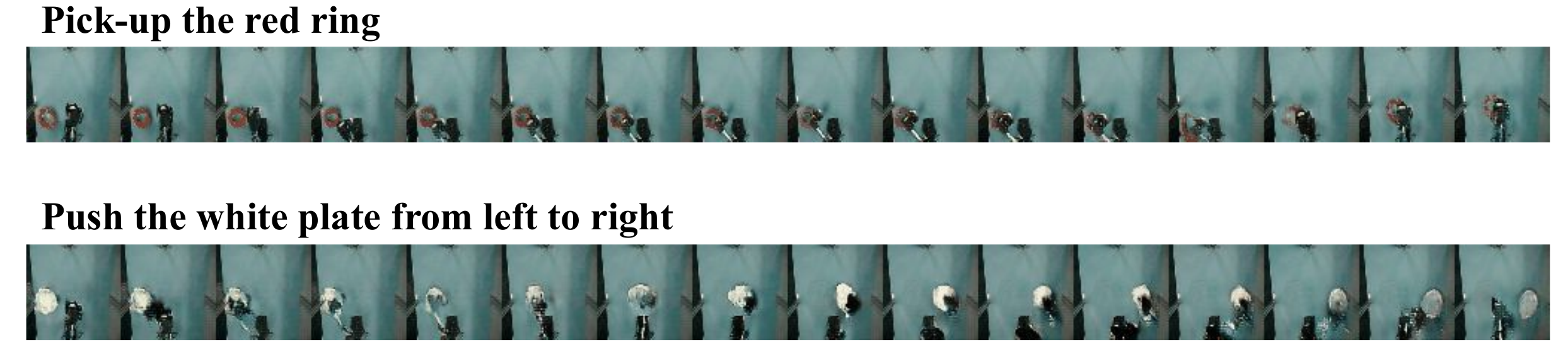}
    \caption{Robotic dataset qualitative results. Corresponding to each user command (sentence) we show the frames of generated videos. All samples are 16 frames with $64 \times 64$ resolution.}
    \label{fig:RobotQual}
\end{figure*}

For both of UCF101, and A2D dataset, we randomly select 5 to 9 frames, with skip rate of 1 frame ($\simeq$ 15 fps); meaning that the training clips can be from  the beginning, middle or end of a video sequence. For the Robotic dataset, we sample 16 frames from a full length demonstration of the robot that can be up to 20 seconds. Thus, the videos shown in Figures~\ref{fig:A2DQualitative}, and ~\ref{fig:UCF101Qualitative} represent about 0.5 of an actual video, and the videos shown in Figure~\ref{fig:RobotQual} covers a longer range (up to 20 seconds) of time. 
We train the models on different datasets in slightly different manners. We use 1 Titan X Pascal GPU to train the experiments of UCF101, and 4 GPUs to train the A2D dataset. Due to the higher variance of videos in A2D dataset, it takes more time for our model to start generating meaningful videos. The model takes 1 day to train on UCF101 and Robotic, and 3 days on A2D.
We employ Spectral Normalization~\cite{miyato2018spectral} on both of Generator and Discriminator modules in all training iterations. We train the Generator and Discriminator equally, i.e., training Generator and Discriminator alternatively, with one iteration for each. We use Adam optimizer with learning rate 0.0001 for both of G and D.

\noindent Please refer to the Supplementary Materials of this manuscript for more Qualitative examples, videos, and etc. We also provide qualitative results for a "Smooth Transition" test. This test is an important evidence of the model generalization. 

{}

\section{Conclusion}
In this paper, we tackle the problem of text to video generation on realistic datasets with free-form sentences. Also, we provide superior results of our proposed Latent Linear Interpolation (LLI) based method compared to well-known approaches. We believe that solving the video content creation using text has a lot of research value and has many real-world usages.

{\small
\bibliographystyle{eccv2020kit/ieee}
\bibliography{egbib}

\begin{thebibliography}{10}\itemsep=-1pt

\bibitem{abolghasemi2019pay}
P.~Abolghasemi, A.~Mazaheri, M.~Shah, and L.~Boloni.
\newblock Pay attention!-robustifying a deep visuomotor policy through
  task-focused visual attention.
\newblock In {\em Proceedings of the IEEE Conference on Computer Vision and
  Pattern Recognition}, pages 4254--4262, 2019.

\bibitem{arik2017deep}
S.~{\"O}. Arik, M.~Chrzanowski, A.~Coates, G.~Diamos, A.~Gibiansky, Y.~Kang,
  X.~Li, J.~Miller, A.~Ng, J.~Raiman, et~al.
\newblock Deep voice: Real-time neural text-to-speech.
\newblock In {\em Proceedings of the 34th International Conference on Machine
  Learning-Volume 70}, pages 195--204. JMLR. org, 2017.

\bibitem{carreira2017quo}
J.~Carreira and A.~Zisserman.
\newblock Quo vadis, action recognition? a new model and the kinetics dataset.
\newblock In {\em proceedings of the IEEE Conference on Computer Vision and
  Pattern Recognition}, pages 6299--6308, 2017.

\bibitem{chan2018everybody}
C.~Chan, S.~Ginosar, T.~Zhou, and A.~A. Efros.
\newblock Everybody dance now.
\newblock {\em arXiv preprint arXiv:1808.07371}, 2018.

\bibitem{clark2019efficient}
A.~Clark, J.~Donahue, and K.~Simonyan.
\newblock Efficient video generation on complex datasets.
\newblock {\em arXiv preprint arXiv:1907.06571}, 2019.

\bibitem{de2017modulating}
H.~De~Vries, F.~Strub, J.~Mary, H.~Larochelle, O.~Pietquin, and A.~C.
  Courville.
\newblock Modulating early visual processing by language.
\newblock In {\em Advances in Neural Information Processing Systems}, pages
  6594--6604, 2017.

\bibitem{deng2009imagenet}
J.~Deng, W.~Dong, R.~Socher, L.-J. Li, K.~Li, and L.~Fei-Fei.
\newblock Imagenet: A large-scale hierarchical image database.
\newblock In {\em 2009 IEEE conference on computer vision and pattern
  recognition}, pages 248--255. Ieee, 2009.

\bibitem{devlin2018bert}
J.~Devlin, M.-W. Chang, K.~Lee, and K.~Toutanova.
\newblock Bert: Pre-training of deep bidirectional transformers for language
  understanding.
\newblock {\em arXiv preprint arXiv:1810.04805}, 2018.

\bibitem{finn2016unsupervised}
C.~Finn, I.~Goodfellow, and S.~Levine.
\newblock Unsupervised learning for physical interaction through video
  prediction.
\newblock In {\em Advances in neural information processing systems}, pages
  64--72, 2016.

\bibitem{fu2019edge}
Z.~Fu, Y.~Zheng, H.~Ye, Y.~Kong, J.~Yang, and L.~He.
\newblock Edge-aware deep image deblurring.
\newblock {\em arXiv preprint arXiv:1907.02282}, 2019.

\bibitem{gavrilyuk2018actor}
K.~Gavrilyuk, A.~Ghodrati, Z.~Li, and C.~G. Snoek.
\newblock Actor and action video segmentation from a sentence.
\newblock In {\em Proceedings of the IEEE Conference on Computer Vision and
  Pattern Recognition}, pages 5958--5966, 2018.

\bibitem{heusel2017gans}
M.~Heusel, H.~Ramsauer, T.~Unterthiner, B.~Nessler, and S.~Hochreiter.
\newblock Gans trained by a two time-scale update rule converge to a local nash
  equilibrium.
\newblock In {\em Advances in Neural Information Processing Systems}, pages
  6626--6637, 2017.

\bibitem{hinz2019semantic}
T.~Hinz, S.~Heinrich, and S.~Wermter.
\newblock Semantic object accuracy for generative text-to-image synthesis.
\newblock {\em arXiv preprint arXiv:1910.13321}, 2019.

\bibitem{huang1996whistler}
X.~Huang, A.~Acero, J.~Adcock, H.-W. Hon, J.~Goldsmith, J.~Liu, and M.~Plumpe.
\newblock Whistler: A trainable text-to-speech system.
\newblock In {\em Proceeding of Fourth International Conference on Spoken
  Language Processing. ICSLP'96}, volume~4, pages 2387--2390. IEEE, 1996.

\bibitem{karras2017progressive}
T.~Karras, T.~Aila, S.~Laine, and J.~Lehtinen.
\newblock Progressive growing of gans for improved quality, stability, and
  variation.
\newblock {\em arXiv preprint arXiv:1710.10196}, 2017.

\bibitem{kay2017kinetics}
W.~Kay, J.~Carreira, K.~Simonyan, B.~Zhang, C.~Hillier, S.~Vijayanarasimhan,
  F.~Viola, T.~Green, T.~Back, P.~Natsev, et~al.
\newblock The kinetics human action video dataset.
\newblock {\em arXiv preprint arXiv:1705.06950}, 2017.

\bibitem{kim2019deep}
D.~Kim, S.~Woo, J.-Y. Lee, and I.~S. Kweon.
\newblock Deep video inpainting.
\newblock In {\em Proceedings of the IEEE Conference on Computer Vision and
  Pattern Recognition}, pages 5792--5801, 2019.

\bibitem{lee2018savp}
A.~X. Lee, R.~Zhang, F.~Ebert, P.~Abbeel, C.~Finn, and S.~Levine.
\newblock Stochastic adversarial video prediction.
\newblock {\em arXiv preprint arXiv:1804.01523}, 2018.

\bibitem{li2018video}
Y.~Li, M.~R. Min, D.~Shen, D.~Carlson, and L.~Carin.
\newblock Video generation from text.
\newblock In {\em Thirty-Second AAAI Conference on Artificial Intelligence},
  2018.

\bibitem{liu2019cross}
Y.~Liu, X.~Wang, Y.~Yuan, and W.~Zhu.
\newblock Cross-modal dual learning for sentence-to-video generation.
\newblock In {\em Proceedings of the 27th ACM International Conference on
  Multimedia}, pages 1239--1247. ACM, 2019.

\bibitem{lotter2016deep}
W.~Lotter, G.~Kreiman, and D.~Cox.
\newblock Deep predictive coding networks for video prediction and unsupervised
  learning.
\newblock {\em arXiv preprint arXiv:1605.08104}, 2016.

\bibitem{Marwah_2017_ICCV}
T.~Marwah, G.~Mittal, and V.~N. Balasubramanian.
\newblock Attentive semantic video generation using captions.
\newblock In {\em The IEEE International Conference on Computer Vision (ICCV)},
  Oct 2017.

\bibitem{mcintosh2018multi}
B.~McIntosh, K.~Duarte, Y.~S. Rawat, and M.~Shah.
\newblock Multi-modal capsule routing for actor and action video segmentation
  conditioned on natural language queries.
\newblock {\em arXiv preprint arXiv:1812.00303}, 2018.

\bibitem{miyato2018spectral}
T.~Miyato, T.~Kataoka, M.~Koyama, and Y.~Yoshida.
\newblock Spectral normalization for generative adversarial networks.
\newblock {\em arXiv preprint arXiv:1802.05957}, 2018.

\bibitem{Pan_2019_CVPR}
J.~Pan, C.~Wang, X.~Jia, J.~Shao, L.~Sheng, J.~Yan, and X.~Wang.
\newblock Video generation from single semantic label map.
\newblock In {\em The IEEE Conference on Computer Vision and Pattern
  Recognition (CVPR)}, June 2019.

\bibitem{pan2017create}
Y.~Pan, Z.~Qiu, T.~Yao, H.~Li, and T.~Mei.
\newblock To create what you tell: Generating videos from captions.
\newblock In {\em Proceedings of the 25th ACM international conference on
  Multimedia}, pages 1789--1798. ACM, 2017.

\bibitem{salimans2016improved}
T.~Salimans, I.~Goodfellow, W.~Zaremba, V.~Cheung, A.~Radford, and X.~Chen.
\newblock Improved techniques for training gans.
\newblock In {\em Advances in neural information processing systems}, pages
  2234--2242, 2016.

\bibitem{schuldt2004recognizing}
C.~Schuldt, I.~Laptev, and B.~Caputo.
\newblock Recognizing human actions: a local svm approach.
\newblock In {\em Proceedings of the 17th International Conference on Pattern
  Recognition, 2004. ICPR 2004.}, volume~3, pages 32--36. IEEE, 2004.

\bibitem{shi2016real}
W.~Shi, J.~Caballero, F.~Husz{\'a}r, J.~Totz, A.~P. Aitken, R.~Bishop,
  D.~Rueckert, and Z.~Wang.
\newblock Real-time single image and video super-resolution using an efficient
  sub-pixel convolutional neural network.
\newblock In {\em Proceedings of the IEEE conference on computer vision and
  pattern recognition}, pages 1874--1883, 2016.

\bibitem{siarohin2019animating}
A.~Siarohin, S.~Lathuili{\`e}re, S.~Tulyakov, E.~Ricci, and N.~Sebe.
\newblock Animating arbitrary objects via deep motion transfer.
\newblock In {\em Proceedings of the IEEE Conference on Computer Vision and
  Pattern Recognition}, pages 2377--2386, 2019.

\bibitem{soomro2012ucf101}
K.~Soomro, A.~R. Zamir, and M.~Shah.
\newblock Ucf101: A dataset of 101 human actions classes from videos in the
  wild.
\newblock {\em arXiv preprint arXiv:1212.0402}, 2012.

\bibitem{spampinato2018vos}
C.~Spampinato, S.~Palazzo, P.~D'Oro, F.~Murabito, D.~Giordano, and M.~Shah.
\newblock Vos-gan: Adversarial learning of visual-temporal dynamics for
  unsupervised dense prediction in videos.
\newblock {\em arXiv preprint arXiv:1803.09092}, 2018.

\bibitem{srivastava2015unsupervised}
N.~Srivastava, E.~Mansimov, and R.~Salakhudinov.
\newblock Unsupervised learning of video representations using lstms.
\newblock In {\em International conference on machine learning}, pages
  843--852, 2015.

\bibitem{tulyakov2018mocogan}
S.~Tulyakov, M.-Y. Liu, X.~Yang, and J.~Kautz.
\newblock Mocogan: Decomposing motion and content for video generation.
\newblock In {\em Proceedings of the IEEE conference on computer vision and
  pattern recognition}, pages 1526--1535, 2018.

\bibitem{villegas2017learning}
R.~Villegas, J.~Yang, Y.~Zou, S.~Sohn, X.~Lin, and H.~Lee.
\newblock Learning to generate long-term future via hierarchical prediction.
\newblock In {\em Proceedings of the 34th International Conference on Machine
  Learning-Volume 70}, pages 3560--3569. JMLR. org, 2017.

\bibitem{wang2018video}
T.-C. Wang, M.-Y. Liu, J.-Y. Zhu, G.~Liu, A.~Tao, J.~Kautz, and B.~Catanzaro.
\newblock Video-to-video synthesis.
\newblock {\em arXiv preprint arXiv:1808.06601}, 2018.

\bibitem{XuHsXiCVPR2015}
C.~Xu, S.-H. Hsieh, C.~Xiong, and J.~J. Corso.
\newblock Can humans fly? {Action} understanding with multiple classes of
  actors.
\newblock In {\em {Proceedings of IEEE Conference on Computer Vision and
  Pattern Recognition}}, 2015.

\bibitem{xu2018attngan}
T.~Xu, P.~Zhang, Q.~Huang, H.~Zhang, Z.~Gan, X.~Huang, and X.~He.
\newblock Attngan: Fine-grained text to image generation with attentional
  generative adversarial networks.
\newblock In {\em Proceedings of the IEEE Conference on Computer Vision and
  Pattern Recognition}, pages 1316--1324, 2018.

\bibitem{yu2018generative}
J.~Yu, Z.~Lin, J.~Yang, X.~Shen, X.~Lu, and T.~S. Huang.
\newblock Generative image inpainting with contextual attention.
\newblock In {\em Proceedings of the IEEE Conference on Computer Vision and
  Pattern Recognition}, pages 5505--5514, 2018.

\end{thebibliography}
}

\end{document}